\documentclass{article}
\usepackage[preprint]{colm2026_conference}

\usepackage{microtype}
\usepackage{hyperref}
\usepackage{url}
\usepackage{booktabs}
\usepackage{amsmath}
\usepackage{amssymb}
\usepackage{graphicx}
\usepackage{multirow}
\usepackage{xspace}
\usepackage{lineno}
\usepackage{enumitem}
\usepackage{algorithm}
\usepackage{algpseudocode}
\usepackage{afterpage}
\usepackage{placeins}
\usepackage{array}

\definecolor{darkblue}{rgb}{0, 0, 0.5}
\hypersetup{colorlinks=true, citecolor=darkblue, linkcolor=darkblue, urlcolor=darkblue}

\newcommand{\selfdoubt}{\textsc{SelfDoubt}\xspace}
\newcommand{\hvr}{\ensuremath{\mathrm{HVR}}\xspace}
\newcommand{\eg}{e.g.,\xspace}

\title{\selfdoubt: Uncertainty Quantification for Reasoning LLMs\\
via the Hedge-to-Verify Ratio}

\author{
Satwik Pandey\thanks{Equal contribution.} \\
Independent Researcher \\
\texttt{psatwik2711@gmail.com}
\And
Suresh Raghu\footnotemark[1] \\
Independent Researcher \\
\texttt{sureshraghu0706@gmail.com}
\And
Shashwat Pandey \\
Zillow Group \\
\texttt{shashwatp@zillowgroup.com}
}

\begin{document}

\ifcolmsubmission
\linenumbers
\fi

\maketitle

\begin{abstract}

Uncertainty estimation for reasoning language models remains difficult to deploy in practice: sampling-based methods are computationally expensive, while common single-pass proxies such as verbalized confidence or trace length are often inconsistent across models. This problem is compounded for proprietary reasoning APIs that expose neither logits nor intermediate token probabilities, leaving practitioners with no reliable uncertainty signal at inference time. We propose \selfdoubt, a single-pass uncertainty framework that resolves this impasse by extracting behavioral signals directly from the reasoning trace itself. Our key signal, the Hedge-to-Verify Ratio (\hvr), detects whether a reasoning trace contains uncertainty markers and, if so, whether they are offset by explicit self-checking behavior. Unlike methods that require multiple sampled traces or model internals, \selfdoubt\ operates on a single observed reasoning trajectory, making it suitable for latency- and cost-constrained deployment over any proprietary API. We evaluate \selfdoubt\ across seven models and three multi-step reasoning benchmarks (BBH, GPQA-Diamond, and MMLU-Pro). Most notably, traces containing no hedging markers are correct 96\% of the time, revealing an emergent high-precision confidence gate at zero additional cost. For the remaining cases, the full \selfdoubt\ score significantly outperforms sampling-based semantic entropy at 10$\times$ lower inference cost. A deployment cascade combining both stages attains 90\% accuracy at 71\% coverage without any task-specific labels. These results establish \selfdoubt\ as a scalable, production-ready foundation for uncertainty estimation over proprietary reasoning models. All code is publicly available at \url{https://github.com/satwik2711/SelfDoubt}.
\end{abstract}

\section{Introduction}
\label{sec:intro}

Uncertainty quantification for reasoning language models remains caught between
two unappealing extremes. Sampling-based methods, including Semantic Entropy
\citep[SE;][]{kuhn2023semantic_uncertainty, farquhar2024detecting_semantic_entropy},
P(True) self-evaluation \citep{kadavath2022mostly_know}, and semantic-geometric
approaches \citep{li2025semantic_volume, phillips2025geometric_uncertainty},
require $N$ independent forward passes, which is prohibitive when a single
reasoning pass already costs thousands of tokens. Single-pass alternatives
exist, but each has a limiting tradeoff: verbalized confidence
\citep{tian2023just_ask_calibration, yoon2025reasoning_models_confidence} is
miscalibrated on hard tasks and weaker models; trace length
\citep{devic2025trace_length} correlates with uncertainty only on
intermediate-difficulty benchmarks; and probe-based methods
\citep{kossen2024sep} require hidden-state access. A further constraint: commercially deployed reasoning APIs do not expose
log-probabilities on reasoning tokens, making methods that depend on token-level
probabilities \citep{moslonka2025epr, zhang2025cot_uq} architecturally
ineligible where UQ is most needed.

We propose \selfdoubt, a single-pass uncertainty framework that extracts
behavioral signals directly from the reasoning trace. Its key signal, the
Hedge-to-Verify Ratio (HVR), is a single scalar: the number of hedge markers
divided by the number of verify markers plus one. Intuitively, hedging
expresses doubt (\textit{``maybe,'' ``perhaps,'' ``not sure''}), whereas
verification acts on it (\textit{``let me check,'' ``verify,'' ``substitute
back''}). HVR therefore captures whether expressed doubt is resolved or left
open. Z-score fusion of HVR with verbalized confidence yields the full
\selfdoubt score, requiring zero training, no model internals, and only 90
unlabeled calibration traces per model.

\begin{enumerate}[leftmargin=1.5em, itemsep=0pt, topsep=2pt]
  \item \textbf{HVR\,=\,0 gate.} We introduce a zero-cost correctness filter
  requiring only regex matching against a marker dictionary. Traces with zero
  hedging language are correct 96.1\% of the time (25.4\% coverage, 0.58\%
  genuine error rate after label-noise correction) across 7 models and 3
  datasets.

  \item \textbf{\selfdoubt.} We present an $O(1)$ uncertainty score fusing HVR
  with verbalized confidence via z-score normalization. \selfdoubt
  significantly outperforms Semantic Entropy on discrimination ($p{=}0.001$)
  while matching its selective prediction quality at $10\times$ lower inference
  cost.

  \item \textbf{Unsupervised marker discovery.} We introduce an unsupervised per-model pipeline that builds
  model-specific dictionaries from 90 unlabeled traces: no correctness labels,
  no manual curation, and no retraining when switching models.

  \item \textbf{Production deferral cascade.} Tier~1 (HVR\,=\,0 gate) plus
  Tier~2 (calibrated z-sum threshold) yields 71\% coverage at 89.7\%
  accuracy, a +9.2~pt lift requiring 4 stored scalars per model.
\end{enumerate}
 
\begin{figure*}[!t]
\centering
\begin{minipage}[t]{0.88\textwidth}
\centering
\includegraphics[width=\linewidth]{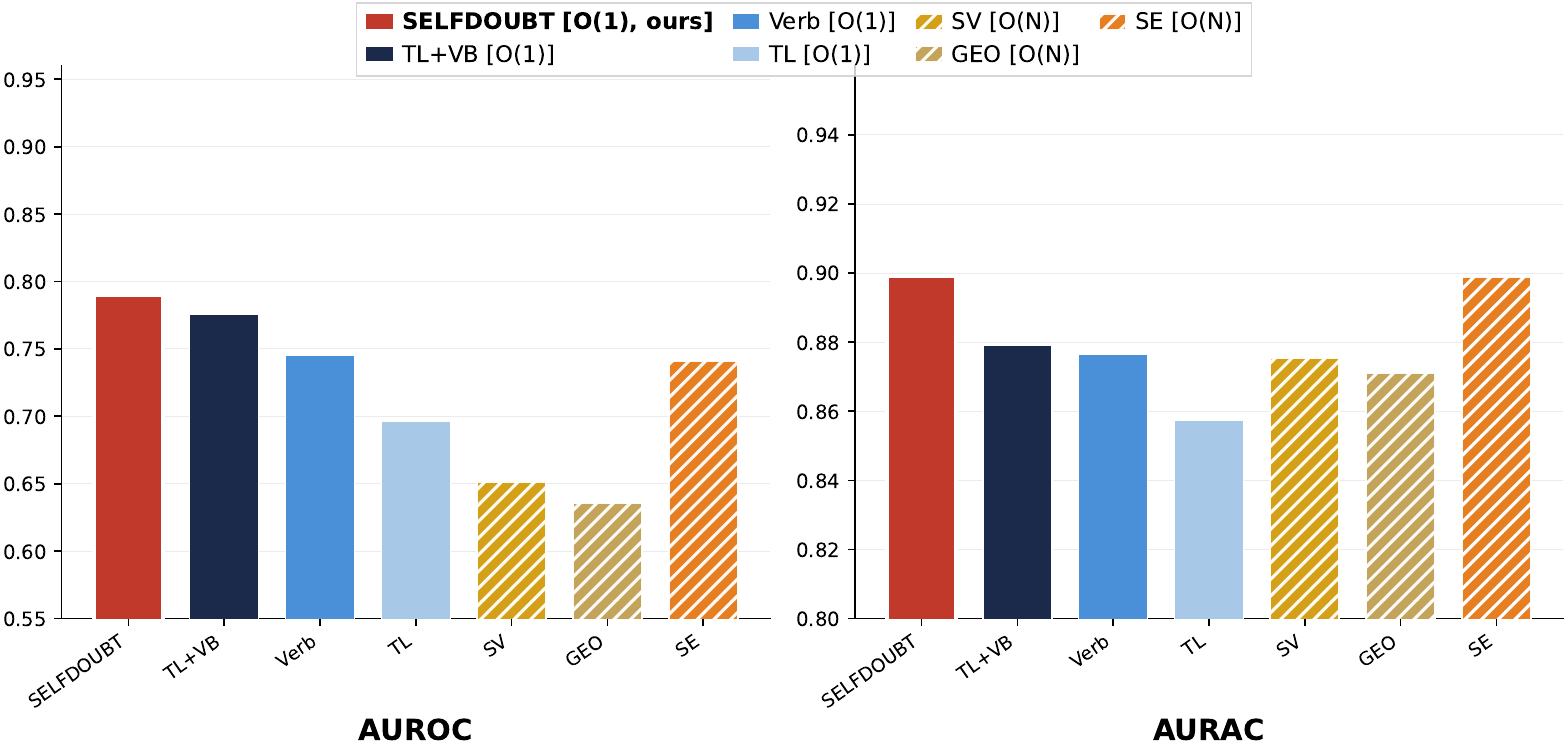}
\vspace{0.2em}
\small \textbf{(a)} Mean AUROC and AURAC across all 21 runs
\end{minipage}
\caption{Headline results. \textbf{(a)} Across 21 runs, \selfdoubt is the
only method that leads both metrics at $O(1)$ cost: mean AUROC 0.7895 and
mean AURAC 0.8992 (slightly above SE at 0.8988).}
\label{fig:headline}
\end{figure*}

\section{Related Work}
\label{sec:related}

\noindent\textbf{Method landscape:}
Table~\ref{tab:landscape} organizes UQ methods along the axes critical for
reasoning-model deployment: computational cost and method requirements.

\begin{table}[t]
\caption{Uncertainty quantification methods for language models. $O(1)$
methods above the first rule; $O(N)$ and $O(N^2)$ below.}
\label{tab:landscape}
\centering
\footnotesize
\setlength{\tabcolsep}{3.5pt}
\begin{tabular}{@{}lcc@{}}
\toprule
Method & Cost & Requirements \\
\midrule
Verb.\ Conf.\ \citep{tian2023just_ask_calibration, yoon2025reasoning_models_confidence} & $O(1)$ & Prompt mod. \\
Trace Length \citep{devic2025trace_length} & $O(1)$ & Trace \\
TL+VB \citep{devic2025trace_length} & $O(1)$ & Trace + prompt mod. \\
Lex.\ Hints \citep{vanhoyweghen2025lexical_hints} & $O(1)$ & Trace + sentiment scorer \\
CoT-UQ \citep{zhang2025cot_uq} & $O(1)$ & Token logits \\
EPR/WEPR \citep{moslonka2025epr} & $O(1)$ & Top-$K$ logprobs \\
SEP \citep{kossen2024sep} & $O(1)$ & Hidden states \\
Latent CoE \citep{wang2024latent_coe} & $O(1)$ & Hidden states \\
\midrule
P(True) self-eval.\ \citep{kadavath2022mostly_know} & $O(N)$ & Self-eval + sampled refs (often logprobs) \\
SE \citep{kuhn2023semantic_uncertainty, farquhar2024detecting_semantic_entropy} & $O(N)$ & $N$ samples + NLI \\
SV \citep{li2025semantic_volume} & $O(N)$ & $N$ samples + log-det(Gram) \\
GEO \citep{phillips2025geometric_uncertainty} & $O(N)$ & $N$ samples + convex hull \\
Topo-UQ \citep{da2025topo_uq} & $O(N^2)$ & $N$ samples + GED \\
\midrule
\textbf{\selfdoubt} & $\mathbf{O(1)}$ & \textbf{Trace + Prompt mod.} \\
\bottomrule
\end{tabular}
\end{table}

\noindent\textbf{Sampling-based UQ:}
Semantic Entropy \citep{kuhn2023semantic_uncertainty,
farquhar2024detecting_semantic_entropy} clusters $N$ sampled outputs by
meaning; Semantic Volume \citep{li2025semantic_volume} and Geometric
Uncertainty \citep{phillips2025geometric_uncertainty} measure embedding spread;
Topo-UQ \citep{da2025topo_uq} computes graph edit distance at $O(N^2)$ cost.
All require 10--20$\times$ inference cost, which is
particularly prohibitive for reasoning models where a
single pass already costs thousands of tokens.

\noindent\textbf{Trace-based $O(1)$ methods:}
\citet{devic2025trace_length} showed trace length is an emergent uncertainty
signal, and TL+VB improves over either component alone.
\citet{yoon2025reasoning_models_confidence} found reasoning models are better
calibrated when verbalizing confidence.
The closest concurrent work is \citet{vanhoyweghen2025lexical_hints}, who
showed hedging words in CoT traces reduce accuracy by up to 40\% relative.
\citet{podolak2025read_your_own_mind} further showed that reasoning traces
surface self-confidence signals absent in direct-answer generations.

\noindent\textbf{Probe-based and hidden-state methods:} Probe-based methods
(SEP \citep{kossen2024sep}, Latent CoE \citep{wang2024latent_coe}) require
hidden states; logprob methods (EPR/WEPR \citep{moslonka2025epr}, CoT-UQ
\citep{zhang2025cot_uq}) require token probabilities unavailable from
production APIs. Both are included in Table~\ref{tab:landscape} but excluded
from comparison; What remains missing is an $O(1)$ method that captures
deliberation structure from text alone, without trained components or model
internals.

\section{\selfdoubt}
\label{sec:method}

\begin{figure}[t]
\centering
\includegraphics[width=\linewidth]{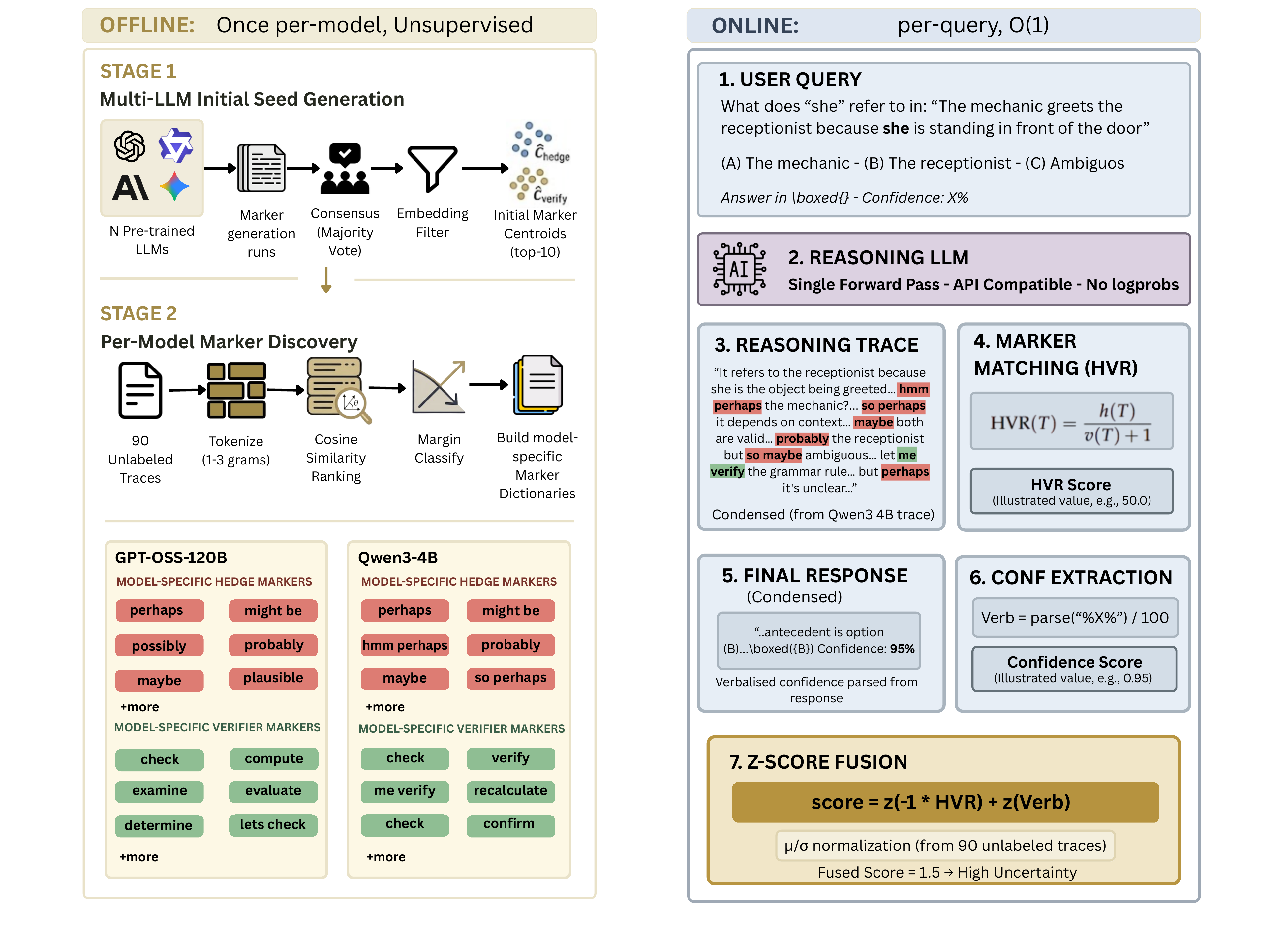}
\caption{\selfdoubt pipeline: calibration-time marker discovery and
inference-time scoring.}
\label{fig:selfdoubt_pipeline}
\end{figure}

Reasoning traces do more than expose a model's final answer: they also
expose how the model expresses and resolves doubt during reasoning. \selfdoubt turns that behavioral
signal into a single-pass uncertainty score. (Figure~\ref{fig:selfdoubt_pipeline}).

\subsection{Data-Driven Marker Discovery}
\label{subsec:discovery}

The central challenge in HVR is not the ratio itself, but defining what should
count as hedging and verification across different models. Reasoning models do
not express uncertainty through a uniform vocabulary, so a fixed lexicon can
miss model-specific phrasing and weaken the signal. We therefore use a
two-stage data-driven pipeline: Stage~1 builds seed vocabularies for hedging
and verification, and Stage~2 expands those seeds into model-specific marker
dictionaries from unlabeled traces.

\textbf{Stage 1: Data-Driven Seed Generation.}
Rather than hand-specifying initial hedge and verify words, we generate them
from model consensus. We query multiple language models to produce candidate
single-word lists for two behaviors: language associated with uncertainty, and
language associated with actively checking or confirming an answer. Each model
is queried five times. We then keep only words that are stable within a model
(appearing in a majority of its runs) and stable across models (appearing in a
majority of models), yielding a consensus candidate set for each role.
Exact prompt templates, model list, and voting rules are given in
\hyperref[sec:appendix-m]{Appendix~M}.

To remove weak or noisy candidates, we run an iterative semantic coherence
filter using \texttt{BAAI/bge-m3} embeddings \citep{chen2024bgem3}; the
pipeline is broadly robust to embedding model choice
(\hyperref[sec:appendix-e]{Appendix~E}). We embed
all candidate words, compute a centroid for the active set, drop outliers below
a cosine-similarity threshold of 0.7 (while preserving at least 10 words),
recompute the centroid, and repeat for up to 6 rounds. The surviving words are
ranked by cosine-to-centroid coherence and materialized into subsets
$\{\text{top\_2}, \ldots, \text{top\_20}\}$. We use \texttt{top\_10} as the
default seed set, selected by the seed-size ablation
(\hyperref[sec:appendix-f]{Appendix~F}). 

\textbf{Stage 2: Per-Model Marker Expansion.}

Given the seed vocabularies from Stage~1, we expand them into model-specific
marker dictionaries using unlabeled reasoning traces. We sample 90 traces per
model across the evaluation datasets and extract candidate 1--3-grams from
their text. To suppress idiosyncratic phrasing, we retain only n-grams that
appear in at least 9 out of the 90 distinct traces (10\%).
\hyperref[sec:appendix-d]{Appendix~D} shows that marker discovery is broadly
stable across calibration set sizes, so we use 90 traces as our standard
setting.

We then embed each candidate with \texttt{bge-m3} and score it by how much
more closely it aligns with the verify centroid than with the hedge centroid:
\[
  \Delta(g) = \cos(g,\, c_{\text{verify}}) - \cos(g,\, c_{\text{hedge}})
\]
A candidate enters the verify dictionary if
\(\Delta(g) > \tau_{\text{verify}}\), enters the hedge dictionary if
\(\Delta(g) < -\tau_{\text{hedge}}\), and is otherwise discarded. This
creates a model-specific neutral band around zero that filters ambiguous or
weakly aligned n-grams. The verify and hedge thresholds are set separately
for each model so that semantically relevant markers are retained while
domain-specific vocabulary is excluded. Exact threshold values are given in
\hyperref[sec:appendix-b]{Appendix~B}; sensitivity to these choices is
analyzed in \hyperref[sec:appendix-n]{Appendix~N}.

The result is a per-model hedge dictionary $\mathcal{H}$ and verify
dictionary $\mathcal{V}$, compiled into word-boundary regex patterns for
trace scanning at inference time. 
Markers transfer across datasets (\hyperref[sec:appendix-g]{Appendix~G}).

\subsection{Hedge-to-Verify Ratio}
\label{subsec:hvr}

Given the per-model dictionaries $\mathcal{H}$ and $\mathcal{V}$ produced
above, we define the signal that consumes them. Let $h(T)$ be the total
number of hedge-marker occurrences in trace $T$ (matched via word-boundary
regex against $\mathcal{H}$, non-overlapping, longest-match-first), and
$v(T)$ the corresponding count against $\mathcal{V}$. The Hedge-to-Verify
Ratio is:

\begin{equation}
  \hvr(T) = \frac{h(T)}{v(T) + 1}
  \label{eq:hvr}
\end{equation}

The $+1$ in the denominator ensures finite values when no verify markers
appear. We orient HVR so that larger values correspond to greater uncertainty; 
empirically, this direction is consistent across all 21 runs.
The distribution is zero-inflated by construction: any trace with zero hedge 
matches yields $\hvr=0$ exactly. We analyze this HVR$=0$ regime separately in
Section~\ref{sec:results} and later operationalize it as a deployment-time
accept rule in Section~\ref{sec:deployment}.

Stage~2 per-model marker expansion is retained as the default because the
HVR$=0$ gate is sensitive to marker coverage
(\hyperref[sec:appendix-h]{Appendix~H}).

\subsection{Verbalized Confidence}
\label{subsec:verb}

In addition to trace-derived uncertainty, we use a direct self-reported signal:
\emph{verbalized confidence}. For each query, the model is prompted to output a
final answer together with a confidence in percent form. We parse this value
from the final answer region and map it to a scalar $V \in [0,1]$ by dividing
the reported percentage by 100, where larger $V$ indicates greater reported
confidence in correctness.

\subsection{\selfdoubt Score}
\label{subsec:score}

HVR and verbalized confidence capture complementary failure modes. HVR captures
uncertainty revealed through deliberation behavior, while verbalized confidence
captures uncertainty the model explicitly reports when asked. We combine them
with additive z-score fusion:

\begin{equation}
  s_{\textsc{sd}}(T) = z_r\!\left(-\hvr(T)\right) + z_r(V),
  \label{eq:selfdoubt}
\end{equation}

where $V$ is the verbalized confidence score and $z_r(x)=(x-\mu_r)/\sigma_r$
standardizes each channel within run $r$ over the usable joined subset. The
negation aligns HVR with correctness, and standardization makes the two channels
commensurable despite model-specific differences in scale.

\section{Experimental Setup}
\label{sec:setup}

\textbf{Models.}
We evaluate seven reasoning models spanning two trace types. Four produce
full reasoning traces via local inference: Qwen3 4B, Qwen3 14B, GPT OSS
20B, and GPT OSS 120B. Three produce compressed thought summaries via
commercial APIs: Claude Sonnet 4.6, Grok 4.1 Fast, and Gemini 2.5 Flash.
Thought summaries expose a provider-compressed version of internal reasoning
rather than the full token-level trace, making uncertainty estimation harder
(Section~\ref{subsec:failures}). Model details are in the respective
technical reports \citep{agarwal2025gpt_oss, yang2025qwen3,
anthropic2026claude_sonnet, xai2025grok, gemini_team2025gemini}.

\textbf{Datasets.}
GPQA-Diamond \citep[198 questions, graduate-level science;][]{rein2024gpqa},
BBH (Big-Bench Hard) \citep[300  stratified questions, diverse
reasoning;][]{suzgun2023bbh}, and MMLU-Pro \citep[300  stratified questions, 10-option
multiple choice;][]{wang2024mmlu_pro}. Together these span three difficulty
profiles and question styles. All datasets are multiple-choice. We evaluated
21 runs: 7 models $\times$ 3 datasets.

\textbf{Baselines.}
All baselines are evaluated on the same samples. $O(1)$ baselines:
Verbalized Confidence (Verb; \citealp{tian2023just_ask_calibration,
yoon2025reasoning_models_confidence}), Trace Length
(TL; \citealp{devic2025trace_length}), and TL+VB
\citep{devic2025trace_length}. $O(N)$ baselines: Semantic Entropy (SE;
\citealp{farquhar2024detecting_semantic_entropy}), Semantic Volume (SV;
\citealp{li2025semantic_volume}), and Geometric Uncertainty (GEO;
\citealp{phillips2025geometric_uncertainty}). We run all sampling-based
baselines with $N{=}10$ samples, incurring roughly $10\times$ the generation
cost of a single-pass approach.
 
\textbf{Metrics.}
\emph{AUROC} (primary): area under the ROC curve measuring
discrimination, i.e., how well the score separates correct from incorrect answers.
\emph{AURAC} (secondary): area under the Risk--Coverage curve measuring
selective prediction quality, i.e., how well the method's ranking supports deferral
policies across all possible thresholds.

\textbf{Statistical testing.}
We use paired Wilcoxon signed-rank tests (one-sided alternative: \selfdoubt
$>$ baseline) on 21 matched model$\times$dataset run pairs.

\begin{table}[htbp]
\caption{Precision and coverage of the HVR\,=\,0 subset with 95\% Wilson
confidence intervals, grouped by trace format. Counts are shown as $N_0 / N$.}
\label{tab:hvr0}
\centering
\small
\setlength{\tabcolsep}{3.5pt}
\begin{tabular}{llrrcl}
\toprule
Group & Model & $N_0/N$ & Coverage &Accuracy (HVR\,=\,0) & Wilson 95\% CI \\
\midrule
\multirow{4}{*}{Full traces}
& Qwen 3 4B & 31/722 & 4.3\% & 90.3\% & [75.1\%, 96.7\%] \\
& Qwen 3 14B & 77/794 & 9.7\% & 96.1\% & [89.2\%, 98.7\%] \\
& GPT OSS 20B & 233/771 & 30.2\% & 96.1\% & [92.8\%, 98.0\%] \\
& GPT OSS 120B & 205/798 & 25.7\% & 96.6\% & [93.1\%, 98.3\%] \\
\midrule
\multirow{3}{*}{Thought summaries}
& Claude Sonnet 4.6 & 425/797 & 53.3\% & 98.1\% & [96.3\%, 99.0\%] \\
& Grok 4.1 Fast & 406/798 & 50.9\% & 94.8\% & [92.2\%, 96.6\%] \\
& Gemini 2.5 Flash & 7/775 & 0.9\% & 57.1\% & [25.1\%, 84.2\%] \\
\midrule
\textbf{All} & \textbf{Pooled (21 runs)} & \textbf{1384/5455} & \textbf{25.4\%} & \textbf{96.1\%} & [94.9\%, 97.0\%] \\
\bottomrule
\end{tabular}
\begin{flushleft}
\end{flushleft}
\end{table}

\section{Results}
\label{sec:results}

\subsection{The Zero-Hedge Regime}
\label{subsec:hvr_gate}

Pooled across 21 runs (5455 queries), traces with $\hvr = 0$ are correct
96.1\% of the time (1330/1384) at 25.4\% coverage
(Table~\ref{tab:hvr0}). The property holds at $\geq$90\% precision for
6 of 7 models; Gemini is the sole outlier, producing only 7 zero-hedge
traces across all three datasets.

Manual error analysis of the 54 raw disagreements reveals that only 8/1384
(0.58\%) represent genuine model failures: the model answered with high
confidence and was factually wrong. The remaining 46 disagreements are grading
artifacts (30 cases: incorrect answer keys or format mismatches) or ambiguous
question labels (16 cases). The label-noise-corrected precision is 99.4\%
(Wilson 95\% CI: [98.9\%, 99.7\%]). Details in
\hyperref[sec:appendix-c]{Appendix~C}.

Thought summary models have higher coverage (35.4\%) than raw-trace models
(17.7\%), reflecting compressed traces that surface fewer hedge markers.
Coverage is highly model-dependent, ranging from 0.9\% (Gemini) to 53.3\%
(Claude).

\FloatBarrier

\subsection{Main Comparison}
\label{subsec:main_results}

Moving from the binary regime to the continuous score,
Table~\ref{tab:main} reports mean AUROC and AURAC across all 21 runs,
grouped by trace format.

\begin{table}[htbp]
\caption{Main results across 21 runs (7 models $\times$ 3 datasets). ``All''
is the mean over all runs; ``Trace'' and ``Summ.'' are means over the 12 full-trace
and 9 thought-summary runs, respectively. $O(1)$ methods above rule, $O(N)$
below. \textbf{Bold}: best; \underline{underline}: second-best.}
\label{tab:main}
\centering
\small
\setlength{\tabcolsep}{3.5pt}
\begin{tabular}{@{}lccccccc@{}}
\toprule
Method & Cost & \multicolumn{2}{c}{All} & \multicolumn{2}{c}{Trace} & \multicolumn{2}{c}{Summ.} \\
\cmidrule(lr){3-4}\cmidrule(lr){5-6}\cmidrule(lr){7-8}
 &  & AUROC & AURAC & AUROC & AURAC & AUROC & AURAC \\
\midrule
\textbf{\selfdoubt (ours)} & $O(1)$ & \textbf{0.7895} & \textbf{0.8992} & \textbf{0.7984} & \textbf{0.8977} & \underline{0.7777} & \underline{0.9013} \\
TL+VB & $O(1)$ & \underline{0.7754} & 0.8792 & 0.7659 & 0.8869 & \textbf{0.7880} & 0.8689 \\
HVR (ours) & $O(1)$ & 0.7509 & 0.8883 & \underline{0.7674} & 0.8884 & 0.7288 & 0.8882 \\
Verb & $O(1)$ & 0.7453 & 0.8765 & 0.7475 & 0.8687 & 0.7424 & 0.8869 \\
TL & $O(1)$ & 0.6959 & 0.8575 & 0.6738 & 0.8644 & 0.7255 & 0.8483 \\
\midrule
SE & $O(N)$ & 0.7407 & \underline{0.8988} & 0.7566 & \underline{0.8916} & 0.7195 & \textbf{0.9085} \\
SV & $O(N)$ & 0.6513 & 0.8753 & 0.6927 & 0.8749 & 0.5961 & 0.8758 \\
GEO & $O(N)$ & 0.6355 & 0.8710 & 0.6619 & 0.8673 & 0.6004 & 0.8759 \\
\bottomrule
\end{tabular}
\end{table}

\selfdoubt achieves the highest mean AUROC (0.7895) and mean AURAC
(0.8992) among all methods at any cost. It is the only $O(1)$ method
competitive with SE on AURAC, matching SE's 0.8988 at a tenth of the
inference cost. Per-dataset, \selfdoubt is strongest on BBH (4/7 AUROC wins),
while GPQA and MMLU-Pro show tighter margins and more competitive baseline
behavior (\hyperref[sec:appendix-a]{Appendix~A}).

\selfdoubt's primary advantage is not per-run dominance but cross-model, cross-metric,
cross-trace-type consistency: best mean on both AUROC and AURAC at $O(1)$ cost, with the
smallest variance across trace formats. This
consistency is what makes it deployable. Section~\ref{sec:deployment} evaluates this directly.

\FloatBarrier

\subsection{Statistical Significance}
\label{subsec:significance}

\begin{table}[t]
\caption{Paired Wilcoxon signed-rank tests (one-sided: \selfdoubt $>$ baseline;
21 matched run pairs). W--D--L = wins--draws--losses for \selfdoubt.}
\label{tab:wilcoxon}
\centering
\small
\begin{tabular}{llrrrcl}
\toprule
Comparison & Metric & Mean $\Delta$ & W--D--L & $W$ & $p$ (one-sided) & Sig.? \\
\midrule
vs.\ SE & AUROC & +0.049 & 17--0--4 & 198 & 0.001 & \textbf{Yes} \\
vs.\ SE & AURAC & +0.000 & 16--0--5 & 159 & 0.069 & No \\
vs.\ Verb & AUROC & +0.044 & 16--1--4 & 190 & $<$0.001 & \textbf{Yes} \\
vs.\ Verb & AURAC & +0.023 & 18--0--3 & 223 & $<$0.001 & \textbf{Yes} \\
vs.\ TL+VB & AUROC & +0.014 & 12--0--9 & 146 & 0.152 & No \\
vs.\ TL+VB & AURAC & +0.020 & 11--0--10 & 142 & 0.187 & No \\
\bottomrule
\end{tabular}
\end{table}

Table~\ref{tab:wilcoxon} reports paired significance tests across 21 matched runs. \selfdoubt significantly outperforms SE on AUROC ($p=0.001$), confirming that an $O(1)$ method can beat the $O(N)$ gold standard at discrimination. The AURAC comparison is not significant ($p=0.069$), indicating equivalent selective prediction quality at one-tenth the cost. \selfdoubt also significantly outperforms Verb-only on both metrics ($p<0.001$), confirming that HVR carries signal beyond verbalized confidence alone. Against TL+VB, results are not significant in either direction. On the full-trace subgroup (12 runs), \selfdoubt significantly beats TL+VB on both metrics ($p=0.026$ AUROC, $p=0.039$ AURAC); on thought summaries, results are competitive but not significant \hyperref[sec:appendix-j]{(Appendix~J)}.

\begin{figure}[t]
\centering
\includegraphics[width=\linewidth]{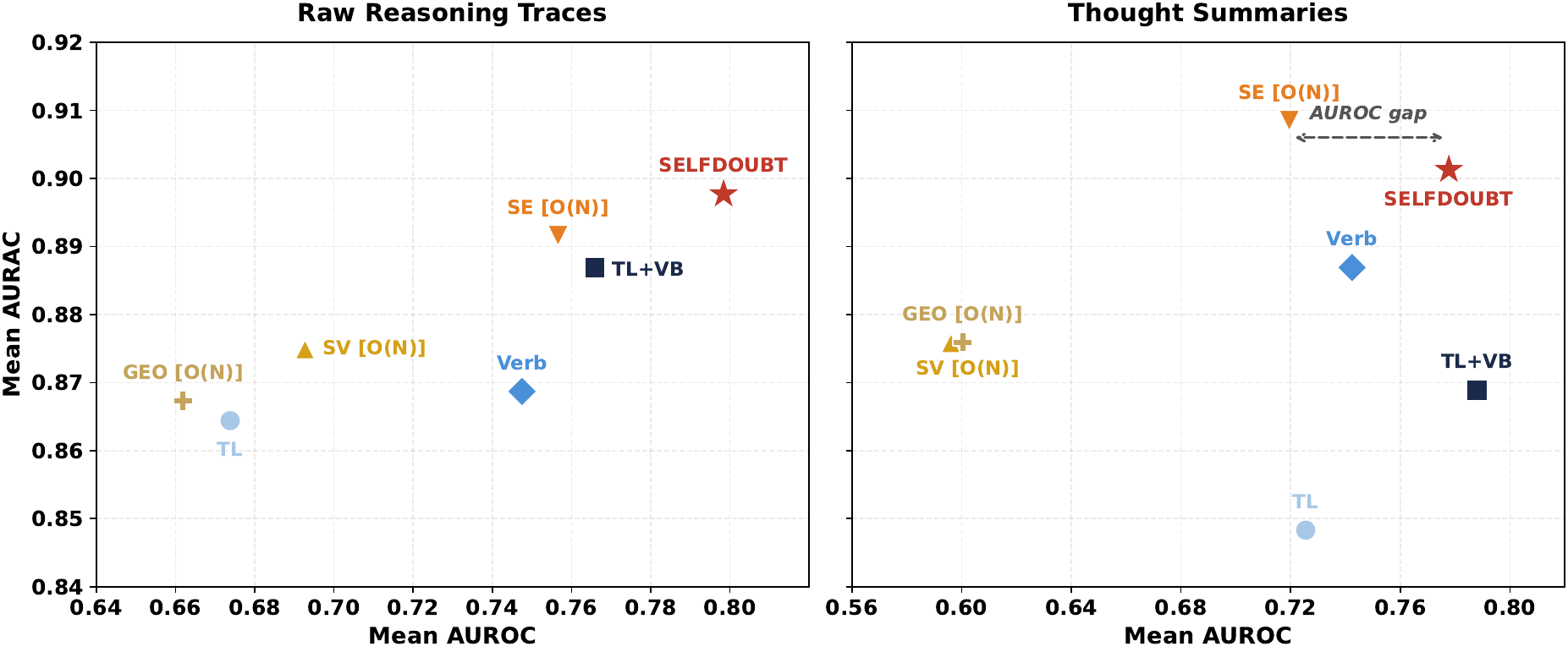}
\caption{AUROC--AURAC consistency on thought summary models. \selfdoubt
occupies the upper-right quadrant among $O(1)$ methods. SE shows strong AURAC
but weak AUROC on compressed traces.}
\label{fig:scatter}
\end{figure}

\subsection{Where \selfdoubt Loses}
\label{subsec:failures}

\selfdoubt does not achieve the best AUROC on the majority of individual
runs. Two failure modes account for most losses. First, when verbalized
confidence is weak on a run (notably Qwen3 4B), adding Verb dilutes HVR
signal. HVR alone outperforms the fused score on these runs. Second,
Gemini's compressed thought summaries produce very sparse n-gram
candidates (0.9\% coverage), making HVR nearly constant and giving trace
length a structural advantage on that model. This suggests a minimum trace richness below which
\selfdoubt cannot operate reliably.

\section{Deployment Policy}
\label{sec:deployment}

A deployable uncertainty method must do more than rank responses: for each
query, it must decide whether to answer or defer. We operationalize \selfdoubt as a
two-stage cascade. Stage~1 auto-accepts traces with zero hedging language (the
HVR\,=\,0 gate). Stage~2 applies calibrated z-score fusion to the remainder,
deferring queries below a tunable threshold.

\textbf{Runtime rule.}
Deployment uses the per-model marker dictionaries from
Section~\ref{sec:method} together with per-model calibration means and standard
deviations for HVR and verbalized confidence, estimated from the 90-trace
calibration set.

\textbf{Tier~1 (HVR\,=\,0 gate).}
The first stage exploits the sharp structural effect identified in
Section~\ref{subsec:hvr_gate}: traces with no hedging language form a distinct
high-precision subset. Operationally, these cases can be accepted immediately,
since detecting them requires only marker matching and no score computation.
To avoid unstable behavior on models where zero-hedge traces are too rare to
calibrate reliably, we enable the gate only when the calibration set contains
at least four such examples ($N_0 \geq 4$); in practice this excludes only
Gemini ($N_0{=}2$). Under this deployed policy, Tier~1 alone answers 25.2\% of
queries at 96.3\% accuracy. (99.4\% after label-noise correction; Section~\ref{subsec:hvr_gate})

\textbf{Tier~2 (calibrated z-sum).} The remaining queries
($\hvr > 0$) are scored by the calibrated fusion
\[
s \;=\; z_{\text{cal}}({-}\hvr) + z_{\text{cal}}(V)
\;=\;
\frac{-\hvr - \mu_{\text{hvr}}}{\sigma_{\text{hvr}}}
+
\frac{V - \mu_{\text{v}}}{\sigma_{\text{v}}},
\]
where $\mu_{\text{hvr}}, \sigma_{\text{hvr}}, \mu_{\text{v}}, \sigma_{\text{v}}$
are estimated from the calibration set, and the query is accepted when
$s \geq \tau$. Because both terms are standardized on that sample,
$\tau{=}0$ is the natural symmetric default: it accepts queries whose fused
evidence is above the calibration mean and defers the rest. More conservative
or more permissive operating points are obtained by shifting $\tau$ upward or
downward according to domain risk tolerance.

\textbf{Calibration cost.}
Moving from oracle normalization (evaluation-set statistics, unavailable at
deployment time) to calibration-set normalization (90-trace statistics) costs
0.48 percentage points of AURAC (0.8992 $\to$ 0.8944). This small gap shows
that z-score fusion transfers from evaluation-time normalization to
deployment-time calibration with minimal loss.

\begin{figure}[!t]
\centering
\includegraphics[width=0.75\linewidth]{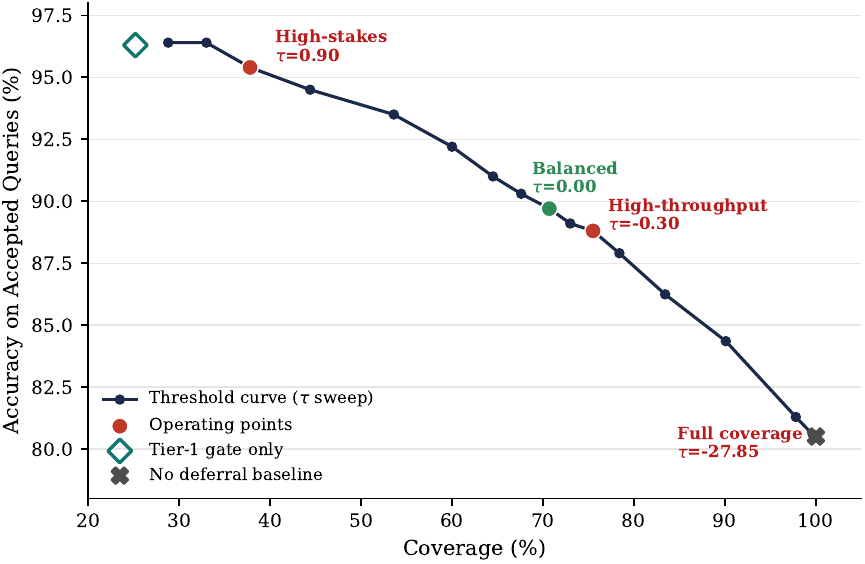}
\caption{Accuracy--coverage curve for the \selfdoubt deployment cascade.}
\label{fig:cascade}
\end{figure}

Figure~\ref{fig:cascade} shows the full accuracy--coverage tradeoff. The
curve is smooth and monotonic, confirming that practitioners can tune $\tau$ to any desired operating point. 
At $\tau{=}0$, the cascade reaches the
balanced operating point emphasized in the paper: 70.7\% coverage at 89.7\%
accuracy. Higher thresholds push the system toward high-stakes use cases;
lower thresholds recover more throughput while degrading gracefully toward the
full-coverage baseline.

The calibrated z-sum remains the strongest Tier-2 ranker we tested: holding Tier 1 fixed, it achieves cascade AURAC 0.8944 and Tier-2 AUROC 0.7572 on the $\hvr > 0$ subset, compared with 0.8830 and 0.7250 for SE \hyperref[sec:appendix-i]{Appendix~I}. Deployment requires only per-model marker dictionaries and four stored scalars, all computed from the same 90 unlabeled traces used for marker discovery.

\section{Limitations}
\label{sec:limitations}

\begin{enumerate}[leftmargin=1.5em, itemsep=0pt, topsep=2pt]

  \item \textbf{HVR\,=\,0 coverage is model-dependent.} 
  For models with very sparse thought
  summaries, the Tier-1 gate provides negligible throughput benefit, and the
  practical value of the deployment cascade is reduced.

  \item \textbf{Style sensitivity.} HVR is defined over surface-form hedging and
  verification markers. A model or prompt that suppresses hedging language, or
  that inserts verification-like phrases without substantive checking, could
  distort both the continuous HVR score and the HVR\,=\,0 gate. Robustness to
  such stylistic drift is untested.

  \item \textbf{No uncertainty decomposition.} \selfdoubt produces a single
  uncertainty score and does not distinguish epistemic from aleatoric sources
  of uncertainty. This is sufficient for selective prediction, which requires
  only a ranking, but limits applicability in settings where the source of
  uncertainty matters.

  \item \textbf{Multiple-choice only.} All three evaluation datasets use
  multiple-choice format. Whether HVR and \selfdoubt generalize to free-form
  generation tasks (open-ended QA, code generation, mathematics with
  step-by-step scoring) is untested.

\end{enumerate}

\section{Conclusion}
\label{sec:conclusion}

Two findings from our evaluation stand out.

First, HVR\,=\,0 defines a high-precision correctness gate: pooled across 21
runs, zero-hedge traces are correct 96.1\% of the time at 25.4\% coverage,
with only 0.58\% genuine errors after label-noise correction. Second,
z-score fusion of HVR with verbalized confidence achieves the best mean AUROC
and AURAC among all $O(1)$ methods. It significantly outperforms the $O(N)$
baseline Semantic Entropy on discrimination ($p{=}0.001$) while matching it on
selective prediction at one-tenth the inference cost. A production cascade
built from these components reaches 71\% coverage at 89.7\% accuracy, a 9.2-point lift over the no-deferral baseline,
requiring no task-specific labels and only 90 unlabeled calibration traces per
model.

Beyond deployment, these results point to a stable textual regularity in
reasoning traces: correctness correlates with how models hedge and self-check,
a pattern consistent across seven models, two trace types, and three datasets.
Whether other trace features, such as revision patterns, confidence trajectories, or
self-correction timing, carry additional signal is an open question as
reasoning models are deployed in high-stakes settings.

\paragraph{Reproducibility Statement.}
\selfdoubt requires no model training, fine-tuning, or access to model
internals. The full pipeline runs from 90 unlabeled reasoning traces per model
using only regex matching and arithmetic. All marker dictionaries, per-model
thresholds, and Stage~2 expansion parameters are provided in Appendices~B
and~M. Prompts for verbalized confidence extraction and Stage~1 seed
generation are reproduced verbatim in Appendix~M. Deployment requires four
stored scalars per model (two means, two standard deviations) estimated from
the same 90-trace calibration set.

\bibliography{references}
\bibliographystyle{colm2026_conference}

\newpage
\appendix

%% ============================================================
\section*{Appendix A: Full Per-Run Results}
\phantomsection\label{sec:appendix-a}
%% ============================================================

\noindent Table~\ref{tab:full_auroc_raw} and Table~\ref{tab:full_auroc_summary} report
per-run AUROC for all 21 runs. Tables~\ref{tab:full_aurac_raw}
and~\ref{tab:full_aurac_summary} report per-run AURAC. Bold indicates per-run
SOTA.
\selfdoubt achieves the highest or second-highest AUROC on the majority of
raw-trace runs. On thought summaries, per-run leadership is split between
\selfdoubt, TL+VB, and SE, consistent with the non-significant pairwise
comparisons reported in Table~\ref{tab:wilcoxon}.

\begin{table}[h]
\caption{Per-run AUROC: raw reasoning trace models.}
\label{tab:full_auroc_raw}
\centering
\small
\begin{tabular}{llccccccccc}
\toprule
Dataset & Model & N & SD & HVR & Verb & TL & TL+VB & SE & SV & GEO \\
\midrule
BBH & gpt\_120b & 300 & \textbf{0.832} & 0.742 & \textbf{0.832} & 0.623 & 0.817 & 0.779 & 0.749 & 0.729 \\
BBH & gpt\_20b & 292 & 0.773 & 0.703 & 0.778 & 0.568 & 0.780 & \textbf{0.819} & 0.776 & 0.784 \\
BBH & qwen3\_4b & 300 & \textbf{0.747} & 0.733 & 0.667 & 0.649 & 0.702 & 0.732 & 0.628 & 0.505 \\
BBH & qwen3\_14b & 300 & \textbf{0.799} & 0.746 & 0.785 & 0.524 & 0.723 & 0.727 & 0.537 & 0.611 \\
GPQA & gpt\_120b & 198 & \textbf{0.844} & 0.813 & 0.835 & 0.761 & 0.839 & 0.760 & 0.758 & 0.697 \\
GPQA & gpt\_20b & 180 & 0.736 & 0.721 & 0.727 & 0.767 & \textbf{0.777} & 0.646 & 0.689 & 0.556 \\
GPQA & qwen3\_4b & 197 & 0.808 & \textbf{0.812} & 0.633 & 0.777 & 0.770 & 0.751 & 0.711 & 0.684 \\
GPQA & qwen3\_14b & 198 & 0.736 & 0.707 & 0.655 & 0.736 & \textbf{0.752} & 0.733 & 0.672 & 0.613 \\
MMLU & gpt\_120b & 300 & 0.847 & 0.804 & \textbf{0.848} & 0.696 & 0.832 & 0.756 & 0.758 & 0.749 \\
MMLU & gpt\_20b & 299 & 0.864 & 0.814 & 0.826 & 0.685 & 0.818 & \textbf{0.873} & 0.804 & 0.711 \\
MMLU & qwen3\_4b & 225 & 0.768 & \textbf{0.802} & 0.639 & 0.667 & 0.655 & 0.707 & 0.666 & 0.643 \\
MMLU & qwen3\_14b & 296 & \textbf{0.828} & 0.814 & 0.747 & 0.631 & 0.728 & 0.797 & 0.566 & 0.662 \\
\bottomrule
\end{tabular}
\end{table}

\begin{table}[h]
\caption{Per-run AUROC: thought summary models.}
\label{tab:full_auroc_summary}
\centering
\small
\begin{tabular}{llccccccccc}
\toprule
Dataset & Model & N & SD & HVR & Verb & TL & TL+VB & SE & SV & GEO \\
\midrule
BBH & claude & 300 & \textbf{0.809} & 0.761 & 0.795 & 0.611 & 0.714 & 0.653 & 0.570 & 0.535 \\
BBH & grok & 300 & 0.826 & 0.771 & 0.821 & 0.723 & \textbf{0.838} & 0.785 & 0.563 & 0.700 \\
BBH & gemini & 293 & 0.640 & 0.645 & 0.532 & \textbf{0.827} & 0.770 & 0.663 & 0.516 & 0.614 \\
GPQA & claude & 198 & 0.787 & 0.686 & 0.807 & 0.772 & \textbf{0.819} & 0.679 & 0.564 & 0.503 \\
GPQA & grok & 198 & 0.831 & 0.791 & 0.801 & 0.809 & \textbf{0.887} & 0.709 & 0.635 & 0.687 \\
GPQA & gemini & 197 & 0.662 & 0.638 & 0.592 & 0.635 & 0.666 & \textbf{0.771} & 0.649 & 0.609 \\
MMLU & claude & 299 & 0.911 & 0.840 & \textbf{0.918} & 0.793 & 0.903 & 0.750 & 0.670 & 0.640 \\
MMLU & grok & 300 & 0.774 & 0.720 & 0.771 & 0.749 & \textbf{0.791} & 0.710 & 0.590 & 0.519 \\
MMLU & gemini & 285 & \textbf{0.759} & 0.707 & 0.645 & 0.612 & 0.704 & 0.756 & 0.608 & 0.596 \\
\bottomrule
\end{tabular}
\end{table}

\begin{table}[h]
\caption{Per-run AURAC: raw reasoning trace models.}
\label{tab:full_aurac_raw}
\centering
\small
\begin{tabular}{llccccccccc}
\toprule
Dataset & Model & N & SD & HVR & Verb & TL & TL+VB & SE & SV & GEO \\
\midrule
BBH & gpt\_120b & 300 & \textbf{0.975} & 0.947 & 0.962 & 0.932 & 0.971 & 0.955 & 0.972 & 0.970 \\
BBH & gpt\_20b & 292 & 0.963 & 0.955 & 0.948 & 0.933 & \textbf{0.966} & 0.947 & 0.945 & 0.946 \\
BBH & qwen3\_4b & 300 & 0.933 & 0.933 & 0.921 & 0.912 & 0.920 & 0.933 & \textbf{0.939} & 0.908 \\
BBH & qwen3\_14b & 300 & \textbf{0.965} & 0.949 & 0.954 & 0.902 & 0.949 & 0.918 & 0.908 & 0.929 \\
GPQA & gpt\_120b & 198 & \textbf{0.882} & 0.871 & 0.877 & 0.834 & 0.879 & 0.878 & 0.881 & 0.864 \\
GPQA & gpt\_20b & 180 & 0.715 & 0.716 & 0.681 & 0.726 & 0.731 & 0.790 & \textbf{0.821} & 0.779 \\
GPQA & qwen3\_4b & 197 & \textbf{0.860} & 0.859 & 0.749 & 0.835 & 0.836 & 0.795 & 0.794 & 0.783 \\
GPQA & qwen3\_14b & 198 & 0.813 & 0.798 & 0.738 & 0.803 & \textbf{0.813} & 0.810 & 0.746 & 0.730 \\
MMLU & gpt\_120b & 300 & 0.935 & 0.917 & \textbf{0.938} & 0.895 & 0.935 & 0.924 & 0.924 & 0.922 \\
MMLU & gpt\_20b & 299 & \textbf{0.941} & 0.924 & 0.931 & 0.889 & 0.929 & 0.929 & 0.884 & 0.862 \\
MMLU & qwen3\_4b & 225 & 0.862 & 0.872 & 0.828 & 0.834 & 0.813 & \textbf{0.892} & 0.839 & 0.832 \\
MMLU & qwen3\_14b & 296 & \textbf{0.930} & 0.920 & 0.896 & 0.878 & 0.901 & 0.928 & 0.845 & 0.883 \\
\bottomrule
\end{tabular}
\end{table}

AURAC is tighter across methods, with per-run margins often below
1 percentage point. \selfdoubt leads most frequently on raw traces; on
thought summaries, TL+VB is the most common per-run leader.

\begin{table}[h]
\caption{Per-run AURAC: thought summary models.}
\label{tab:full_aurac_summary}
\centering
\small
\begin{tabular}{llccccccccc}
\toprule
Dataset & Model & N & SD & HVR & Verb & TL & TL+VB & SE & SV & GEO \\
\midrule
BBH & claude & 300 & 0.953 & 0.948 & 0.949 & 0.955 & \textbf{0.967} & 0.916 & 0.941 & 0.938 \\
BBH & grok & 300 & 0.970 & 0.960 & 0.964 & 0.957 & \textbf{0.972} & 0.934 & 0.902 & 0.945 \\
BBH & gemini & 293 & 0.709 & 0.711 & 0.634 & 0.371 & 0.371 & 0.904 & 0.906 & \textbf{0.936} \\
GPQA & claude & 198 & 0.925 & 0.881 & 0.928 & 0.913 & \textbf{0.936} & 0.916 & 0.905 & 0.876 \\
GPQA & grok & 198 & 0.947 & 0.937 & 0.938 & 0.925 & \textbf{0.960} & 0.929 & 0.772 & 0.908 \\
GPQA & gemini & 197 & 0.790 & 0.785 & 0.771 & 0.785 & 0.792 & \textbf{0.864} & 0.820 & 0.803 \\
MMLU & claude & 299 & 0.977 & 0.966 & \textbf{0.978} & 0.954 & 0.976 & 0.898 & 0.882 & 0.871 \\
MMLU & grok & 300 & 0.900 & 0.878 & 0.899 & 0.904 & \textbf{0.917} & 0.884 & 0.853 & 0.780 \\
MMLU & gemini & 285 & \textbf{0.940} & 0.928 & 0.921 & 0.871 & 0.930 & 0.931 & 0.902 & 0.825 \\
\bottomrule
\end{tabular}
\end{table}

\FloatBarrier

%% ============================================================
\section*{Appendix B: Marker Discovery Details}
\phantomsection\label{sec:appendix-b}
%% ============================================================

\paragraph{Stage 1 centroid filtering parameters.}
Embedding model: \texttt{BAAI/bge-m3} \citep{chen2024bgem3}; similarity
threshold: 0.7; minimum keep count: 10; maximum iterations: 6. Observed:
hedge candidates remained stable across iterations (no drops); verifier
candidates had noisy outliers removed (\eg \textit{``hmm,''}
\textit{``plugging''}).
Thresholds were selected per model by inspecting margin distributions and
choosing cutoffs that preserve semantically coherent markers while excluding
domain-specific vocabulary; sensitivity to this choice is reported in
\hyperref[sec:appendix-n]{Appendix~N}.

\paragraph{Per-model Stage~2 thresholds.}

\begin{table}[h]
\centering
\small
\caption{Per-model Stage~2 thresholds for marker classification.}
\label{tab:appendix-b-stage2-thresholds}
\begin{tabular}{lcc}
\toprule
Model & $\tau_{\text{verify}}$ & $\tau_{\text{hedge}}$ \\
\midrule
Claude Sonnet 4.6 & 0.09 & 0.08 \\
Gemini 2.5 Flash & 0.12 & 0.05 \\
\texttt{gpt-oss-20b} & 0.14 & 0.20 \\
\texttt{gpt-oss-120b} & 0.10 & 0.15 \\
Grok 4.1 Fast & 0.10 & 0.10 \\
Qwen3 & 0.17 & 0.20 \\
Qwen3-14B & 0.15 & 0.15 \\
\bottomrule
\end{tabular}
\end{table}

%% ============================================================
\section*{Appendix C: Audit of HVR=0 Raw Disagreements}
\phantomsection\label{sec:appendix-c}
%% ============================================================

This appendix details the manual audit underlying the HVR\,=\,0 analysis.
Across the 21 pooled runs (7 models $\times$ 3 datasets), the evaluation set
contains 5455 total rows, of which 1384 have $\hvr=0$ (25.4\% coverage).
Under the original benchmark labels, 1330 of these 1384 rows are marked
correct, giving a raw precision of 96.10\% and leaving 54 apparent
false accepts for audit.

\paragraph{What was audited.}
We manually re-examined \emph{all 54} HVR\,=\,0 rows that were marked incorrect
by the original benchmark labels. The goal of this audit was not to relabel the
entire benchmark, but to determine whether each apparent HVR\,=\,0 failure
represented a genuine confident model error.

\paragraph{Audit categories and decision rules.}
Upon review each of the 54 cases was assigned to exactly one of the following mutually
exclusive categories, using the rules below \emph{in order}:

\begin{enumerate}[leftmargin=1.5em, itemsep=1pt, topsep=2pt]
  \item \textbf{Grading or answer-key issue.}
  The model answer is semantically correct, or at least as defensible as the
  key, but is scored as incorrect because of a benchmark artifact. This includes
  exact-match formatting mismatches (e.g., \texttt{false} vs.\ \texttt{False}),
  option-letter vs.\ full-option-text mismatches, duplicate or semantically
  equivalent answer options, and incorrect answer keys.
  \emph{Decision rule:} if the model answer is independently verifiable as
  correct or equivalent, assign this category.

  \item \textbf{Ambiguous question or ambiguous label.}
  The prompt admits multiple defensible interpretations, or the benchmark label
  draws a boundary that is not uniquely determined by the problem statement.
  \emph{Decision rule:} if a reasonable expert could defend both the model
  answer and the benchmark answer, assign this category.

  \item \textbf{Genuinely wrong confident prediction.}
  The model answer is clearly inconsistent with the problem or gold option, and
  no grading artifact or ambiguity explains the discrepancy.
  \emph{Decision rule:} assign this category only after (1) and (2) are ruled
  out. This is the residual category.
\end{enumerate}

\paragraph{Headline audit result.}
Table~\ref{tab:hvr0_audit_summary} shows the pooled result. Of the 54 apparent
HVR\,=\,0 errors, 46 do not survive audit as genuine false accepts: 30 are
grading or answer-key issues and 16 are ambiguous items. Only 8 cases remain
as genuinely wrong confident predictions. Under this audit, the estimated
false-accept rate of the HVR\,=\,0 gate is therefore 8/1384 = 0.58\%, rather
than the raw 54/1384 = 3.90\%.

\begin{table}[h]
\caption{Audit of all 54 raw HVR\,=\,0 disagreements.}
\label{tab:hvr0_audit_summary}
\centering
\small
\begin{tabular}{lrrr}
\toprule
Variant / category & Count & \% of 54 & \% of $N_0{=}1384$ \\
\midrule
Raw disagreements & 54 & 100.0\% & 3.90\% \\
\quad Grading or answer-key issue & 30 & 55.6\% & 2.17\% \\
\quad Ambiguous question / label & 16 & 29.6\% & 1.16\% \\
\quad Genuinely wrong confident & 8 & 14.8\% & 0.58\% \\
\midrule
Raw precision & 1330 / 1384 & --- & 96.10\% \\
Audit-adjusted precision & 1376 / 1384 & --- & 99.42\% \\
\bottomrule
\end{tabular}
\end{table}

The corresponding Wilson intervals are:
raw precision $= 96.10\%$ (95\% CI: [94.94\%, 97.00\%]) and
audit-adjusted precision $= 99.42\%$ (95\% CI: [98.86\%, 99.71\%]).

\paragraph{Breakdown by dataset and model.}
The 54 raw disagreements are not concentrated in a single benchmark.
By dataset, the audit yields 23 BBH cases, 6 GPQA cases, and 25 MMLU-Pro
cases. By final category, BBH contributes 7 grading/key issues, 12 ambiguous
items, and 4 genuine errors; GPQA contributes 5 grading/key issues, 0
ambiguous items, and 1 genuine error; MMLU-Pro contributes 18 grading/key
issues, 4 ambiguous items, and 3 genuine errors.
By model, the corresponding counts are: Claude (3 / 4 / 1), Gemini (3 / 0 / 0),
GPT-OSS-120B (4 / 3 / 0), GPT-OSS-20B (5 / 2 / 2), Grok (11 / 5 / 5),
Qwen3 (3 / 0 / 0), and Qwen3-14B (1 / 2 / 0), where each triple is
(grading/key, ambiguous, genuine).

\paragraph{Representative audited cases.}
The grading/key bucket is not dominated by borderline rescues; many are
straightforward evaluation artifacts. Three representative examples are
independently checkable.

First, in \texttt{mmlu\_pro.gpt\_oss\_120b.11998}, the model selects option J
(\texttt{+j50 ohms}) while the key marks option E
(\texttt{-j50 ohms}). For a short-circuited transmission-line setting with
$\beta d = \pi$, the standard relation
$Z_{\text{in}} = j Z_0 \tan(\beta d)$ supports the model-side sign convention
used in the worked solution rather than the provided key. Second, in
\texttt{mmlu\_pro.gpt\_oss\_120b.11904}, the model selects option C
(\textit{period}) while the key marks option I
(\textit{row number on periodic table}); these are semantically equivalent.
Third, several cases are pure grading-format failures, including exact-match
mismatches such as \texttt{false} vs.\ \texttt{False} on BBH Boolean
Expressions and option-letter vs.\ full-option-text mismatches.

The ambiguous bucket mainly contains tasks whose label boundary is itself
underspecified. A representative example is BBH Salient Translation Error
Detection, where the distinction between a named-entity error and a factual
error can be label-dependent even when the model's explanation is coherent.

The genuine-error bucket is small but real. The 8 surviving cases have mean
verbalized confidence 0.978. They include three Dyck-language completions with
an extra leading bracket, one clear output-format failure
(\texttt{\textbackslash boxed\{54\}} instead of an option letter), and four
standard wrong-option selections with high stated confidence.

\paragraph{Interpretation.}
The main claim of the paper does not depend on the audit: the raw benchmark
result is already 96.10\% precision at 25.4\% coverage. The audit sharpens the
interpretation of the residual 3.90\% raw error rate by showing that most of it
comes from benchmark artifacts or intrinsically ambiguous labels rather than
from genuinely wrong confident predictions. For this reason, we report both
numbers in the paper: the raw benchmark precision as the primary result, and
the audit-adjusted 99.42\% figure as a secondary diagnostic estimate under the
adjudication criteria above.

\paragraph{Scope, limitations, and sensitivity.}
This audit is a manual post-hoc analysis of the 54 raw disagreements only,
conducted by a single annotator. It should therefore be read as a transparency
measure on the failure modes of the HVR\,=\,0 gate, not as a replacement
benchmark label set. We do not use the audit anywhere in model selection,
thresholding, or score construction. Two structural safeguards partially
mitigate subjectivity: the ordered decision rules ensure that
category~(3) is the residual after (1) and (2) are ruled out, so any
classification ambiguity \emph{inflates} the genuine-error count; and the
30 grading/key cases are independently verifiable against domain references
without annotator judgment.

The main locus of subjectivity is the 16-item ambiguous bucket. To bound the
effect of this judgment call, Table~\ref{tab:hvr0_sensitivity} reports three
scenarios: the audit as annotated; a stricter setting in which all ambiguous
cases are reclassified as genuine errors; and the zero-correction baseline
using the raw labels only. Even under the stricter scenario, precision remains
98.27\% (1360/1384), showing that the central claim does not hinge on trusting
the annotator's judgment on every ambiguous item.

\begin{table}[h]
\caption{Sensitivity of HVR\,=\,0 precision to audit trust level.}
\label{tab:hvr0_sensitivity}
\centering
\small
\begin{tabular}{lrrr}
\toprule
Scenario & Genuine errors & Precision & Wilson 95\% CI \\
\midrule
As annotated & 8 & 99.42\% & [98.86\%, 99.71\%] \\
All ambiguous $\rightarrow$ genuine & 24 & 98.27\% & [97.32\%, 98.89\%] \\
Zero correction (raw labels only) & 54 & 96.10\% & [94.94\%, 97.00\%] \\
\bottomrule
\end{tabular}
\end{table}

%% ============================================================
\section*{Appendix D: Sample Size Ablation}
\phantomsection\label{sec:appendix-d}
%% ============================================================

Performance plateaus near 60 traces for both models; we use 90 as the default
to provide margin against sampling noise.

\paragraph{Stratified sampling.}
\begin{table}[h]
\centering
\small
\caption{Sample-size ablation under stratified sampling.}
\label{tab:appendix-d-stratified}
\begin{tabular}{rrccc}
\toprule
Traces & Sampling & Model & SD AUROC & HVR AUROC \\
\midrule
15 & stratified & Claude & --- (empty markers) & --- \\
30 & stratified & Claude & 0.821 & 0.671 \\
60 & stratified & Claude & 0.829 & 0.751 \\
90 & stratified & Claude & 0.833 & 0.756 \\
120 & stratified & Claude & 0.831 & 0.748 \\
180 & stratified & Claude & 0.827 & 0.743 \\
\midrule
15 & stratified & Qwen3 & 0.779 & 0.783 \\
30 & stratified & Qwen3 & 0.776 & 0.784 \\
60 & stratified & Qwen3 & 0.776 & 0.783 \\
90 & stratified & Qwen3 & 0.774 & 0.782 \\
120 & stratified & Qwen3 & 0.775 & 0.783 \\
180 & stratified & Qwen3 & 0.774 & 0.784 \\
\bottomrule
\end{tabular}
\end{table}

\paragraph{Random pooled sampling.}
\begin{table}[h]
\centering
\small
\caption{Sample-size ablation under random pooled sampling.}
\label{tab:appendix-d-random-pooled}
\begin{tabular}{rrccc}
\toprule
Traces & Sampling & Model & SD AUROC & HVR AUROC \\
\midrule
15 & random\_pooled & Claude & --- & --- \\
30 & random\_pooled & Claude & --- & --- \\
60 & random\_pooled & Claude & 0.835 & 0.754 \\
90 & random\_pooled & Claude & 0.831 & 0.751 \\
120 & random\_pooled & Claude & 0.833 & 0.759 \\
180 & random\_pooled & Claude & 0.828 & 0.746 \\
\midrule
15 & random\_pooled & Qwen3 & 0.776 & 0.779 \\
30 & random\_pooled & Qwen3 & 0.775 & 0.779 \\
60 & random\_pooled & Qwen3 & 0.776 & 0.785 \\
90 & random\_pooled & Qwen3 & 0.776 & 0.784 \\
120 & random\_pooled & Qwen3 & 0.777 & 0.785 \\
180 & random\_pooled & Qwen3 & 0.774 & 0.783 \\
\bottomrule
\end{tabular}
\end{table}

Values are mean SD/HVR AUROC over BBH/GPQA/MMLU-Pro.

%% ============================================================
\section*{Appendix E: Embedding Model Ablation}
\phantomsection\label{sec:appendix-e}
%% ============================================================

\begin{table}[h]
\centering
\small
\caption{Embedding-model ablation for \selfdoubt AUROC across datasets.}
\label{tab:appendix-e-embedding-ablation}
\begin{tabular}{llrrrr}
\toprule
Embedding & Model & BBH & GPQA & MMLU & Mean \\
\midrule
\multicolumn{6}{l}{\textit{SELFDOUBT AUROC}} \\
bge-m3 & Claude & 0.825 & 0.824 & 0.904 & 0.851 \\
OpenAI-3-large & Claude & 0.820 & 0.832 & 0.905 & 0.852 \\
Qwen3-emb-8b & Claude & 0.822 & 0.829 & 0.903 & 0.852 \\
bge-m3 & Qwen3 & 0.748 & 0.793 & 0.758 & 0.766 \\
OpenAI-3-large & Qwen3 & 0.763 & 0.813 & 0.765 & 0.780 \\
Qwen3-emb-8b & Qwen3 & 0.750 & 0.811 & 0.782 & 0.781 \\
\bottomrule
\end{tabular}
\end{table}

SELFDOUBT range: 0.001 (Claude), 0.015 (Qwen3). The maximum spread across
embedding backends is 1.5 percentage points, confirming that marker discovery
is near-immune to encoder choice. Given this robustness, we use
\texttt{bge-m3} as the practical default because it is locally runnable,
inexpensive, and empirically competitive.

%% ============================================================
\section*{Appendix F: Seed Set Size Ablation}
\phantomsection\label{sec:appendix-f}
%% ============================================================

\begin{table}[h]
\centering
\small
\caption{Seed-set size ablation on mean \selfdoubt AUROC.}
\label{tab:appendix-f-seed-size}
\begin{tabular}{lrr}
\toprule
Seed Set & Mean SD AUROC & $\Delta$ vs top\_10 \\
\midrule
top\_10 & \textbf{0.7895} & 0.000 \\
top\_8 & 0.7861 & $-$0.003 \\
top\_20 & 0.7826 & $-$0.007 \\
top\_2 & 0.7822 & $-$0.007 \\
top\_4 & 0.7815 & $-$0.008 \\
top\_12 & 0.7814 & $-$0.008 \\
top\_6 & 0.7813 & $-$0.008 \\
top\_16 & 0.7803 & $-$0.009 \\
random\_10 & 0.7782 & $-$0.011 \\
\bottomrule
\end{tabular}
\end{table}

Performance is stable across coherence-ranked subsets (\texttt{top\_2} through
\texttt{top\_20}), with a maximum spread of 0.8~pt. The gap to
\texttt{random\_10} (1.1~pt) confirms that coherence ranking contributes
meaningful signal, but the method is not brittle to seed-set size within the
ranked family. We select \texttt{top\_10} as the default because it achieves
the highest mean AUROC, while \texttt{top\_8} performs comparably.

%% ============================================================
\section*{Appendix G: Cross-Dataset Marker Transfer (MuSR)}
\phantomsection\label{sec:appendix-g}
%% ============================================================

Per-model means at 90 MuSR
\citep{sprague2024musr} traces:

\begin{table}[h]
\centering
\small
\caption{Cross-dataset marker transfer: per-model mean \selfdoubt AUROC at MuSR@90.}
\label{tab:appendix-g-transfer-musr}
\begin{tabular}{lrrl}
\toprule
Model & Original & MuSR@90 & $\Delta$ \\
\midrule
\texttt{gpt-oss-120b} & 0.8406 & 0.8366 & $-$0.004 \\
\texttt{gpt-oss-20b} & 0.7908 & 0.7867 & $-$0.004 \\
Qwen3 & 0.7745 & 0.7654 & $-$0.009 \\
Qwen3-14B & 0.7876 & 0.7853 & $-$0.002 \\
Claude Sonnet 4.6 & 0.8358 & 0.8358 & 0.000 \\
Grok 4.1 Fast & 0.8101 & 0.8089 & $-$0.001 \\
Gemini 2.5 Flash & 0.6871 & 0.6475 & $-$0.040 \\
\bottomrule
\end{tabular}
\end{table}

\noindent Gemini's degradation is driven by BBH ($-$11.8~pt) while GPQA and
MMLU-Pro are near-flat, consistent with its anomalous sparse-trace behavior.
Excluding Gemini, mean degradation is 0.3 percentage points, confirming that
the transferred markers capture model-specific language more than
dataset-specific vocabulary.

%% ============================================================
\section*{Appendix H: Stage 2 vs.\ Seeds-Only}
\phantomsection\label{sec:appendix-h}
%% ============================================================

For HVR specifically, Stage~2 prevents large
degradations on thought summary models:

\begin{table}[h]
\centering
\small
\caption{Stage~2 data-driven markers vs.\ seeds-only markers for HVR AUROC on summaries.}
\label{tab:appendix-h-stage2-vs-seeds}
\begin{tabular}{llrrl}
\toprule
Model & Dataset & Data-Driven & Seeds Only & $\Delta$ (HVR) \\
\midrule
Claude & BBH & 0.733 & 0.659 & $-$0.074 \\
Claude & MMLU-Pro & 0.834 & 0.700 & $\mathbf{-0.134}$ \\
Grok & MMLU-Pro & 0.720 & 0.678 & $-$0.042 \\
\bottomrule
\end{tabular}
\end{table}

SELFDOUBT AUROC shows a dead split (9--9 wins, +0.005 mean for seeds-only),
indicating that z-score fusion absorbs moderate marker noise. The design
implication is asymmetric: Stage~2 is mandatory for standalone HVR and the
HVR\,=\,0 gate, where marker precision directly controls false-accept risk, but
optional for the fused \selfdoubt score, where confidence fusion compensates.

%% ============================================================
\section*{Appendix I: Deployment Tier-2 Ranker Comparison}
\phantomsection\label{sec:appendix-i}
%% ============================================================

\begin{table}[h]
\centering
\small
\caption{Deployment Tier-2 ranker comparison on hard-subset ranking and cascade performance.}
\label{tab:appendix-i-tier2-ranker}
\begin{tabular}{lrrl}
\toprule
Tier-2 Ranker & Cascade AURAC & T2 AUROC (HVR$>$0) & Cost \\
\midrule
\textbf{z-sum (SD)} & \textbf{0.8944} & \textbf{0.7572} & $O(1)$ \\
HVR only & 0.8883 & 0.6975 & $O(1)$ \\
SE & 0.8830 & 0.7250 & $O(N)$ \\
Verb only & 0.8756 & 0.7180 & $O(1)$ \\
TL+VB & 0.8730 & 0.7149 & $O(1)$ \\
Random & 0.8245 & 0.4917 & --- \\
\bottomrule
\end{tabular}
\end{table}

Z-sum beats SE as Tier-2 ranker by 3.2~pt T2 AUROC and 1.1~pt cascade AURAC
at $O(1)$ cost. HVR-only's cascade AURAC (0.8883) is propped up by the
Tier-1 gate; its T2 AUROC on the hard subset (0.6975) is weakest among
non-random rankers, validating the fusion design.

%% ============================================================
\section*{Appendix J: Subgroup Wilcoxon Tables}
\phantomsection\label{sec:appendix-j}
%% ============================================================

On full traces, \selfdoubt significantly beats all three comparators on AUROC.
On thought summaries, it loses to TL+VB (3--6 W--L) but remains significant
against SE and Verb; this subgroup is where the aggregate TL+VB
non-significance originates.

\begin{table}[h]
\centering
\small
\caption{Subgroup paired Wilcoxon results for \selfdoubt vs.\ baselines on AUROC and AURAC.}
\label{tab:appendix-j-subgroup-wilcoxon}
\begin{tabular}{llrrrl}
\toprule
Group & Comparison & $n$ & Mean $\Delta$ AUROC & W--D--L & $p$ \\
\midrule
Full traces & vs TL+VB & 12 & +0.032 & 9--0--3 & \textbf{0.026} \\
Full traces & vs SE & 12 & +0.042 & 10--0--2 & \textbf{0.005} \\
Full traces & vs Verb & 12 & +0.051 & 9--1--2 & \textbf{0.002} \\
Summaries & vs TL+VB & 9 & $-$0.010 & 3--0--6 & 0.787 \\
Summaries & vs SE & 9 & +0.058 & 7--0--2 & \textbf{0.037} \\
Summaries & vs Verb & 9 & +0.035 & 7--0--2 & \textbf{0.049} \\
\midrule
Full traces & vs TL+VB & 12 & +0.011 AURAC & 8--0--4 & \textbf{0.039} \\
Full traces & vs SE & 12 & +0.006 AURAC & 9--0--3 & 0.102 \\
Full traces & vs Verb & 12 & +0.029 AURAC & 11--0--1 & $<$\textbf{0.001} \\
Summaries & vs TL+VB & 9 & +0.032 AURAC & 3--0--6 & 0.850 \\
Summaries & vs SE & 9 & $-$0.007 AURAC & 7--0--2 & 0.248 \\
Summaries & vs Verb & 9 & +0.014 AURAC & 7--0--2 & \textbf{0.014} \\
\bottomrule
\end{tabular}
\end{table}

%% ============================================================
\section*{Appendix K: Implementation and Hardware Details}
\phantomsection\label{sec:appendix-k}
%% ============================================================

\paragraph{Calibration pass.}
We use 90 traces per model, sampled randomly from the pooled BBH/GPQA/MMLU-Pro
calibration set. Usable matched rows in the current artifacts are:
Claude=90, \texttt{gpt-oss-20b}=86, \texttt{gpt-oss-120b}=90, Qwen3=80,
Qwen3-14B=90, Grok=90, and Gemini=87.

\paragraph{Inference.}
Marker matching uses pre-compiled word-boundary regex patterns
(\texttt{\textbackslash b...\textbackslash b}, case-insensitive). HVR and
z-sum computation require only arithmetic per query (4 stored scalars per
model). Verbalized confidence is extracted from model output via a standard
prompt suffix: ``Confidence: [0-100]\%'' and then rescaled to $[0,1]$.

\paragraph{Model serving.}
Qwen3-4B was run on a rented RTX~3090 GPU. All other closed-model evaluations
were run through OpenRouter, which provides a unified API with multi-provider
routing and automatic fallback across available providers
\citep{openrouter_quickstart,openrouter_provider_routing}. OpenRouter exposes
\texttt{logprobs} and \texttt{top\_logprobs} as optional parameters, but
parameter support is provider-dependent and must be checked at the
model/provider level \citep{openrouter_parameters}. Because token-level
log-probabilities were not available consistently for the OpenRouter-served
model/provider combinations used in our runs, we do not report logit-based
baselines.

\paragraph{Datasets.}
GPQA-Diamond \citep[198 questions;][]{rein2024gpqa},
BBH/Big-Bench Hard \citep[300 stratified questions;][]{suzgun2023bbh},
MMLU-Pro \citep[300 stratified questions;][]{wang2024mmlu_pro}.

%% ============================================================
\section*{Appendix L: AUROC and AURAC Scaling Plots}
\phantomsection\label{sec:appendix-l}
%% ============================================================

\begin{figure}[htbp]
\centering
\begin{minipage}[t]{0.49\linewidth}
\centering
\includegraphics[width=\linewidth]{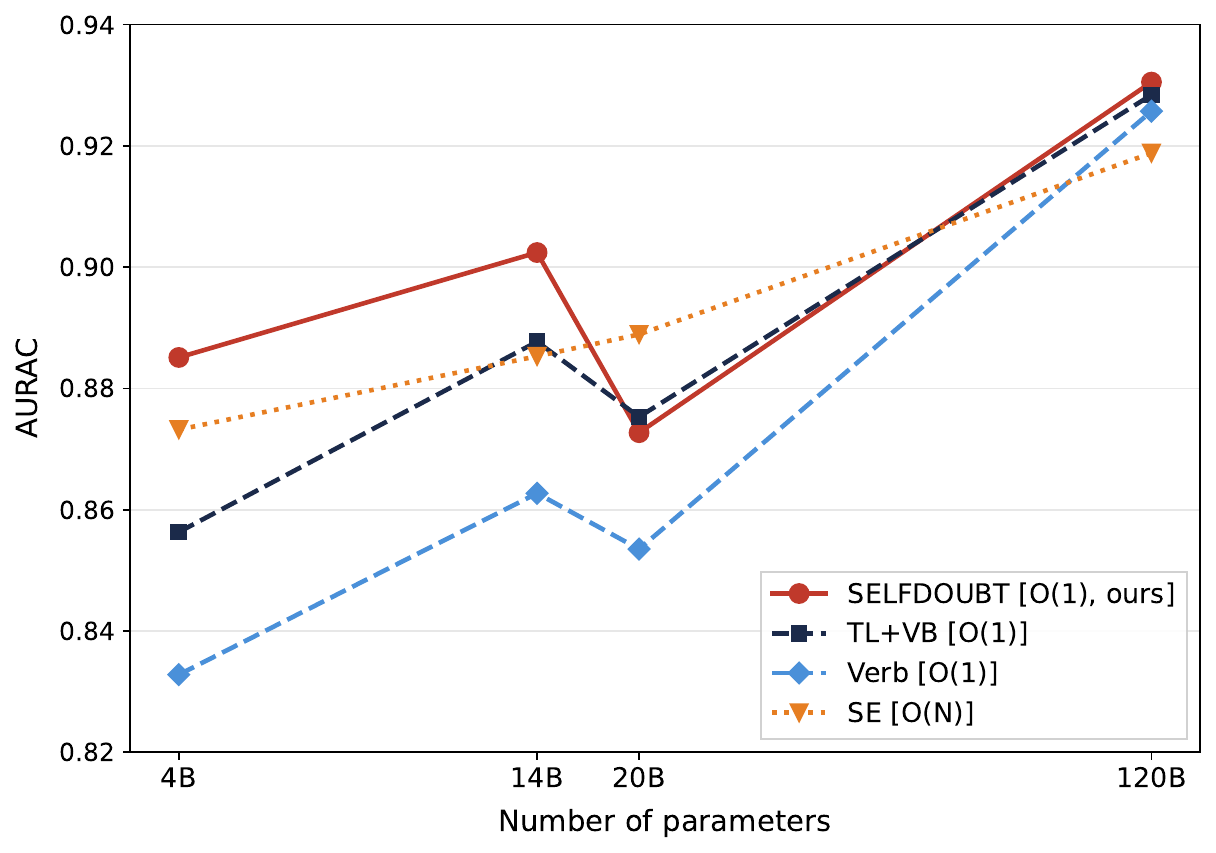}
\small \textbf{(a)} AUROC
\end{minipage}\hfill
\begin{minipage}[t]{0.49\linewidth}
\centering
\includegraphics[width=\linewidth]{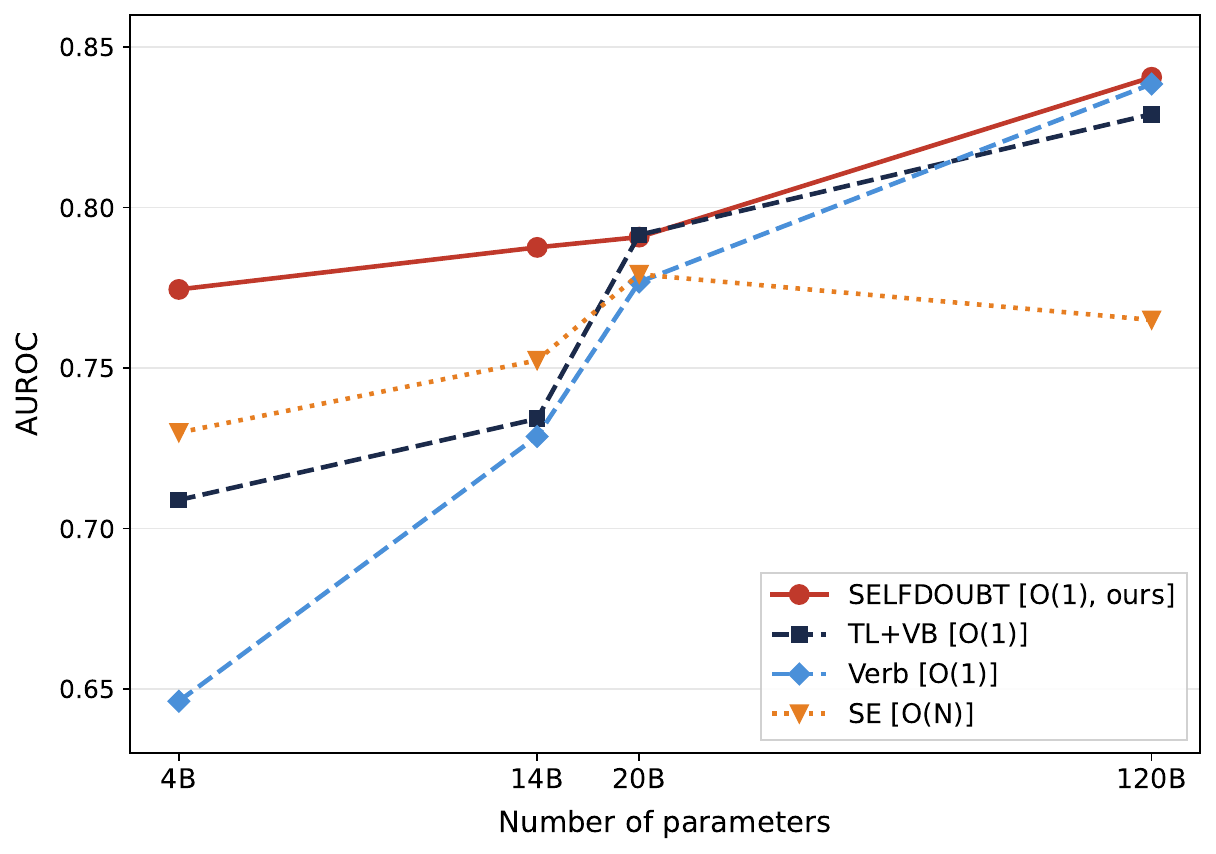}
\small \textbf{(b)} AURAC
\end{minipage}
\caption{Scaling vs.\ model size (raw traces only). \textbf{(a)} AUROC.
\textbf{(b)} AURAC.}
\label{fig:appendix-scaling}
\end{figure}

\noindent Figure~\ref{fig:appendix-scaling} reports AUROC (left) and AURAC
(right) as functions of model size on raw traces only. Across the four plotted
methods, AUROC generally improves with scale, with the clearest monotonic
trend coming from \selfdoubt, which rises from 0.7745 (Qwen3-4B) to 0.8406
(GPT OSS 120B).

\noindent Relative to the baselines, \selfdoubt remains competitive at every
scale on both metrics. TL+VB and Verb generally improve with scale but start
noticeably lower at small models, while SE is less monotonic across the same
range. AURAC is somewhat noisier at intermediate scales, but \selfdoubt still
attains the strongest endpoint at 120B (0.9305).

\FloatBarrier

%% ============================================================
\section*{Appendix M: Prompting and Extraction Details}
\phantomsection\label{sec:appendix-m}
%% ============================================================

\paragraph{M.1 Verbalized-confidence prompting.}
For verbalized-confidence runs, the effective prompt consists of the question
text followed by a task-specific instruction block, separated by two newlines.
Across all datasets, the confidence suffix is standardized as:
\begin{quote}
\small\ttfamily\raggedright
Confidence: [0-100]\%
\end{quote}

\noindent\textbf{GPQA-Diamond.}
\begin{quote}
\small\ttfamily\raggedright
After solving, put only the correct option in the box, and output the final
answer exactly once in the form \textbackslash boxed\{...\}, followed by your
confidence that the answer is correct as a percentage. Use this exact format:
Confidence: [0-100]\%
\end{quote}

\noindent\textbf{MMLU-Pro.}
\begin{quote}
\small\ttfamily\raggedright
After solving, output your final answer exactly once as the option label in
parentheses, e.g., (C). Then output your confidence using this exact format:
Confidence: [0-100]\%
\end{quote}

\noindent\textbf{BBH task families.}
Rather than repeating 27 near-identical prompt strings, we group BBH tasks by
their required answer format. In every case, the answer instruction below is
followed by the same confidence suffix
(\texttt{Confidence: [0-100]\%}).

\begin{table}[h]
\centering
\small
\caption{BBH task-family answer-format instructions used in verbalized-confidence prompting.}
\label{tab:appendix-m-bbh-answer-format}
\setlength{\tabcolsep}{5pt}
\begin{tabular}{>{\raggedright\arraybackslash}p{0.28\linewidth} >{\raggedright\arraybackslash}p{0.64\linewidth}}
\toprule
\textbf{Answer type} & \textbf{Tasks and instruction} \\
\midrule

Boxed \texttt{True}/\texttt{False} &
\texttt{boolean\_expressions}: output the final answer exactly once in the form
\texttt{\textbackslash boxed\{...\}} using only \texttt{True} or
\texttt{False}. \\

Boxed \texttt{Yes}/\texttt{No} &
\texttt{causal\_judgement}, \texttt{navigate}, \texttt{web\_of\_lies}:
output the final answer exactly once in the form
\texttt{\textbackslash boxed\{...\}} using only \texttt{Yes} or \texttt{No}. \\

Boxed lowercase \texttt{yes}/\texttt{no} &
\texttt{sports\_understanding}: output the final answer exactly once in the
form \texttt{\textbackslash boxed\{...\}} using only \texttt{yes} or
\texttt{no}. \\

Boxed \texttt{valid}/\texttt{invalid} &
\texttt{formal\_fallacies}: output the final answer exactly once in the form
\texttt{\textbackslash boxed\{...\}} using only \texttt{valid} or
\texttt{invalid}. \\

Boxed option label &
\texttt{date\_understanding}, \texttt{disambiguation\_qa},
\texttt{geometric\_shapes}, \texttt{hyperbaton},
\texttt{logical\_deduction\_three\_objects},
\texttt{logical\_deduction\_five\_objects},
\texttt{logical\_deduction\_seven\_objects},
\texttt{movie\_recommendation}, \texttt{penguins\_in\_a\_table},
\texttt{reasoning\_about\_colored\_objects}, \texttt{ruin\_names},
\texttt{salient\_translation\_error\_detection}, \texttt{snarks},
\texttt{temporal\_sequences},
\texttt{tracking\_shuffled\_objects\_three\_objects},
\texttt{tracking\_shuffled\_objects\_five\_objects},
\texttt{tracking\_shuffled\_objects\_seven\_objects}: output the final answer
exactly once in the form \texttt{\textbackslash boxed\{...\}} using only the
option label in parentheses, e.g., \texttt{(C)}. \\

Boxed integer &
\texttt{multistep\_arithmetic\_two}: output the final answer exactly once in
the form \texttt{\textbackslash boxed\{...\}} using only the final integer
(with sign if needed). \newline
\texttt{object\_counting}: output the final answer exactly once in the form
\texttt{\textbackslash boxed\{...\}} using only the final integer. \\

Structured free-form output &
\texttt{dyck\_languages}: output the final answer exactly once as
\texttt{Final Answer: <tokens>}, using only the missing bracket tokens
separated by single spaces. \newline
\texttt{word\_sorting}: output the final answer exactly once in the form
\texttt{\textbackslash boxed\{...\}}, using only the sorted words separated by
single spaces. \\

\bottomrule
\end{tabular}
\end{table}

\paragraph{M.2 Confidence parsing.}
Confidence is extracted from the answer region, defined as the text after the
last \texttt{</think>} tag when such a tag is present, and otherwise from the
full response. We first apply a strict parser that looks for an explicit
\texttt{Confidence:} or \texttt{Confidence=} field followed by a numeric
percentage. If no such field is found, we apply a fallback parser that scans
the same answer region for percentage expressions more broadly. In both cases,
we retain the last valid percentage in the range $[0,100]$ and rescale it to
$[0,1]$.

\paragraph{M.3 Stage-1 seed-generation prompts.}
Stage~1 uses one prompt for hedging and one for verification.

\noindent\textbf{Hedge prompt.}
\begin{quote}
\small\ttfamily\raggedright
List 50 single English words that a speaker uses when they are uncertain or
lack confidence in a claim they are making. Return only the words, one per
line, no numbering.
\end{quote}

\noindent\textbf{Verify prompt.}
\begin{quote}
\small\ttfamily\raggedright
List 50 single English words that a speaker uses when they are performing a
concrete action to test or confirm an answer they already gave -- such as
recalculating a value, substituting it back into an equation, cross-checking
against a known fact, or computing an independent check. These words should
indicate the speaker is actively testing correctness, not merely reconsidering
or reflecting. Return only the words, one per line, no numbering.
\end{quote}
 
\paragraph{M.4 Stage-1 generation models and voting rules.}
We query four models (\texttt{gpt-oss-20b}, \texttt{gpt-oss-120b},
\texttt{qwen3-14b}, \texttt{grok-4.1-fast}), requesting five independent runs
per role.

Selection uses a strict-majority rule at two levels:
\begin{enumerate}[leftmargin=1.5em, itemsep=1pt, topsep=2pt]
  \item \textbf{Within model:} keep a word if it appears in at least 3 of 5 runs.
  \item \textbf{Across models:} keep a word if it is retained by at least 3 of 4 models.
\end{enumerate}

The resulting candidate sets are then passed through iterative coherence
pruning using \texttt{BAAI/bge-m3} embeddings, cosine threshold 0.7,
minimum keep count 10, and at most 6 pruning rounds. In the saved artifacts,
the hedge role yields 41 global-majority candidates and retains all 41 after
coherence filtering; the verify role yields 48 global-majority candidates, of
which 46 survive after pruning.

\paragraph{M.5 Example retained seeds.}
Table~\ref{tab:seed-examples} shows the top-10 retained words from the
post-pruning coherence ranking for each role.

\begin{table}[h]
\centering
\small
\caption{Example retained Stage-1 seeds after coherence filtering.}
\label{tab:seed-examples}
\begin{tabular}{p{0.42\linewidth} p{0.42\linewidth}}
\toprule
\textbf{Hedge} & \textbf{Verify} \\
\midrule
possibly & check \\
seemingly & reassess \\
maybe & reevaluate \\
apparently & re-evaluate \\
probably & reinspect \\
presumably & rechecking \\
perhaps & verifying \\
likely & reconfirming \\
reportedly & recheck \\
seems & prove \\
\bottomrule
\end{tabular}
\end{table}

%% ============================================================
\section*{Appendix N: Threshold Sensitivity}
\phantomsection\label{sec:appendix-n}
%% ============================================================

Both Stage~2 classification thresholds ($\tau_{\text{verify}}$ and
$\tau_{\text{hedge}}$) are scaled jointly by a single multiplier applied
to the per-model defaults in Appendix~B. We sweep five multipliers
$\{0.5\times, 0.75\times, 1.0\times, 1.25\times, 1.5\times\}$ and report
mean \selfdoubt AUROC across all 21 runs.

\begin{table}[h]
\centering
\small
\caption{Sensitivity of mean \selfdoubt AUROC to joint Stage~2 threshold scaling.}
\label{tab:appendix-n-threshold-sensitivity}
\begin{tabular}{lllrr}
\toprule
Multiplier & $\tau_{\text{verify}}$ range & $\tau_{\text{hedge}}$ range
  & Mean SD AUROC & $\Delta$ vs.\ 1.0$\times$ \\
\midrule
$0.5\times$  & [0.045, 0.085] & [0.025, 0.100] & 0.7675 & $-$0.022 \\
$0.75\times$ & [0.068, 0.128] & [0.038, 0.150] & 0.7807 & $-$0.009 \\
$\mathbf{1.0\times}$ (default) & [0.090, 0.170] & [0.050, 0.200] & \textbf{0.7895} & 0.000 \\
$1.25\times$ & [0.113, 0.213] & [0.063, 0.250] & 0.7854 & $-$0.004 \\
$1.5\times$  & [0.135, 0.255] & [0.075, 0.300] & 0.7649 & $-$0.025 \\
\bottomrule
\end{tabular}
\end{table}

\noindent Ranges reflect per-model variation from Appendix~B; all models'
thresholds are scaled by the same multiplier simultaneously. The total spread
across all five conditions is 2.5~pt (0.7649--0.7895). Across the central
three conditions ($0.75\times$--$1.25\times$), the spread narrows to 0.9~pt,
confirming that \selfdoubt is robust to threshold choice within a broad
neighborhood of the defaults. Degradation at the extremes ($0.5\times$
and $1.5\times$) is driven primarily by Qwen3, whose marker sets become
either too inclusive or too restrictive when thresholds deviate by 50\%.
The z-score fusion absorbs moderate threshold perturbations: the
\selfdoubt AUROC range (2.5~pt) is substantially smaller than the HVR-alone
range (12.6~pt) across the same conditions, confirming that the fusion
design reduces sensitivity to the heuristic threshold selection.

%% ============================================================
\section*{Appendix O: Algorithmic Summary}
\phantomsection\label{sec:appendix-o}
%% ============================================================

\noindent This appendix gives compact pseudocode for the two implementation
views used throughout the paper: \selfdoubt calibration/scoring and the
deployment-time accept/defer cascade.

\begin{algorithm}[htbp]
\small
\caption{\selfdoubt: Calibration and Scoring}
\label{alg:selfdoubt}
\begin{algorithmic}[1]
\Statex \textbf{Calibration} (once per model)
\Require 90 unlabeled traces $\{T_1, \ldots, T_{90}\}$ with verbalized confidences $\{V_1, \ldots, V_{90}\}$
\Ensure Dictionaries $\mathcal{H}, \mathcal{V}$; scalars $\mu_{\text{hvr}}, \sigma_{\text{hvr}}, \mu_{\text{v}}, \sigma_{\text{v}}$
\Statex \textit{Stage 1: Seed generation}
\State Query 4 LLMs $\times$ 5 runs each for hedge/verify word lists
\State Retain words appearing in a majority of runs and a majority of models
\State Embed candidates with \texttt{bge-m3}; iteratively drop outliers below cosine 0.7 to the centroid (up to 6 rounds, floor of 10 words)
\State Select \texttt{top\_10} ranked seeds $\to$ centroids $c_{\text{hedge}}, c_{\text{verify}}$
\Statex \textit{Stage 2: Per-model marker expansion}
\State Extract 1--3-gram candidates from $\{T_i\}$; retain n-grams seen in at least $\text{min\_trace\_count}$ traces
\State Classify by margin $\Delta(g) = \cos(g, c_{\text{verify}}) - \cos(g, c_{\text{hedge}})$ into $\mathcal{H}$ or $\mathcal{V}$
\Statex \textit{Z-score calibration}
\For{each $T_i$}
    \State $h_i \gets$ hedge occurrences in $T_i$ matched against $\mathcal{H}$
    \State $v_i \gets$ verify occurrences in $T_i$ matched against $\mathcal{V}$
    \State $\text{hvr}_i \gets h_i / (v_i + 1)$
\EndFor
\State $\mu_{\text{hvr}}, \sigma_{\text{hvr}} \gets \text{mean}, \text{std}(\{\text{hvr}_i\})$
\State $\mu_{\text{v}}, \sigma_{\text{v}} \gets \text{mean}, \text{std}(\{V_i\})$
\Statex
\Statex \textbf{Scoring} (per query)
\Require Trace $T$, verbalized confidence $V$
\Ensure \selfdoubt score $s$
\State $h \gets$ hedge occurrences in $T$ matched against $\mathcal{H}$
\State $v \gets$ verify occurrences in $T$ matched against $\mathcal{V}$
\State $\text{hvr} \gets h / (v + 1)$
\State $s \gets \dfrac{-\text{hvr} - \mu_{\text{hvr}}}{\sigma_{\text{hvr}}} + \dfrac{V - \mu_{\text{v}}}{\sigma_{\text{v}}}$
\State \Return $s$
\end{algorithmic}
\end{algorithm}

\begin{algorithm}[htbp]
\small
\caption{SelfDoubt Deployment Cascade (per query)}
\label{alg:inference}
\begin{algorithmic}[1]
\Require Trace $T$, verbalized confidence $V$, dictionaries $\mathcal{H}, \mathcal{V}$, scalars $\mu_{\text{hvr}}, \sigma_{\text{hvr}}, \mu_{\text{v}}, \sigma_{\text{v}}$, threshold $\tau$
\Ensure Decision $\in \{\textsc{accept}, \textsc{defer}\}$
\State $h \gets$ number of word-boundary regex matches from $\mathcal{H}$ in $T$
\State $v \gets$ number of word-boundary regex matches from $\mathcal{V}$ in $T$
\If{$h = 0$} \Comment{Tier 1: HVR\,=\,0 gate}
    \State \Return \textsc{accept}
\EndIf
\State $\text{hvr} \gets h / (v + 1)$ \Comment{Tier 2: scored ranking}
\State $s \gets \dfrac{-\text{hvr} - \mu_{\text{hvr}}}{\sigma_{\text{hvr}}} + \dfrac{V - \mu_{\text{v}}}{\sigma_{\text{v}}}$
\If{$s \geq \tau$}
    \State \Return \textsc{accept}
\Else
    \State \Return \textsc{defer}
\EndIf
\end{algorithmic}
\end{algorithm}

\FloatBarrier

\end{document}